\def\year{2018}\relax
\documentclass[letterpaper]{article} 
\usepackage{aaai18}  
\usepackage{times}  
\usepackage{helvet}  
\usepackage{courier}  
\usepackage{url}  
\usepackage{graphicx}  
\usepackage{latexsym}
\usepackage{subcaption}
\usepackage{booktabs}
\usepackage{arydshln}
\usepackage[]{titlesec}
\begin{document}
%
\title{Production Ready Chatbots: Generate if not Retrieve}
\author{Aniruddha Tammewar, Monik Pamecha, Chirag Jain, Apurva Nagvenkar, Krupal Modi\\Haptik Inc\\
Mumbai, India\\\{anutammewar,apurv.nagvenkar\}@gmail.com, monik@berkeley.edu, \{chirag.jain,krupal\}@haptik.co
}
\maketitle	
\begin{abstract}
In this paper, we present a hybrid model that combines a neural conversational model and a rule-based graph dialogue system that assists users in scheduling reminders through a chat conversation. The graph based system has high precision and provides a grammatically accurate response but has a low recall. The neural conversation model can cater to a variety of requests, as it generates the responses word by word as opposed to using canned responses. The hybrid system shows significant improvements over the existing baseline system of rule based approach and caters to complex queries with a domain-restricted neural model. Restricting the conversation topic and combination of graph based retrieval system with a neural generative model makes the final system robust enough for a real world application. 
\end{abstract}
\section{Introduction}
\label{sect:intro}
Chat interfaces have gained popularity due to the ubiquity of smartphones and connectivity \cite{wildml}. This has lead to an influx of human-to-human conversation data that has motivated the research to design chatbots that can mimic human ability to converse. There has been growing research interest in building such models, which are either domain restricted or open domain \cite{wang2011improving,ameixa2014luke,li2016diversity}. Rule based retrieval models provide well formed responses but suffer from low coverage due to the complex nature of natural language. For example, chatbots like A.L.I.C.E use Artificial Intelligence Markup Language (AIML) \cite{wallace2003elements}, that stores query-response pairs and compares the input to its respository to determine a response. This is essentially a more sophisticated form of hard-coding. The approach is inefficient in responding to different representations of the same query. Such methods are hard to scale and fail to harness the abundant conversation data available. Recent advancements in neural models have brought generative approaches to the forefront because they provide a variety of responses to complicated inputs \cite{sordoni-EtAl:2015:NAACL-HLT,DBLP:journals/corr/VinyalsL15,li2016diversity,ShangLL15}, although such techniques are more prone to grammatical errors.

In this work, we present a robust system that combines the generative and the retrieval approach. We address requests to schedule and cancel reminders. The rule-based graph dialogue model works well for the expected chat flows. The generative model is helpful in keeping the users from deviating from the expected conversational flow by intercepting at turns and directing the user to respond in accordance with the desired flow.

\begin{figure}[!hb]
\begin{subfigure}{.4\columnwidth}
  \centering
  \includegraphics[width=0.8\columnwidth]{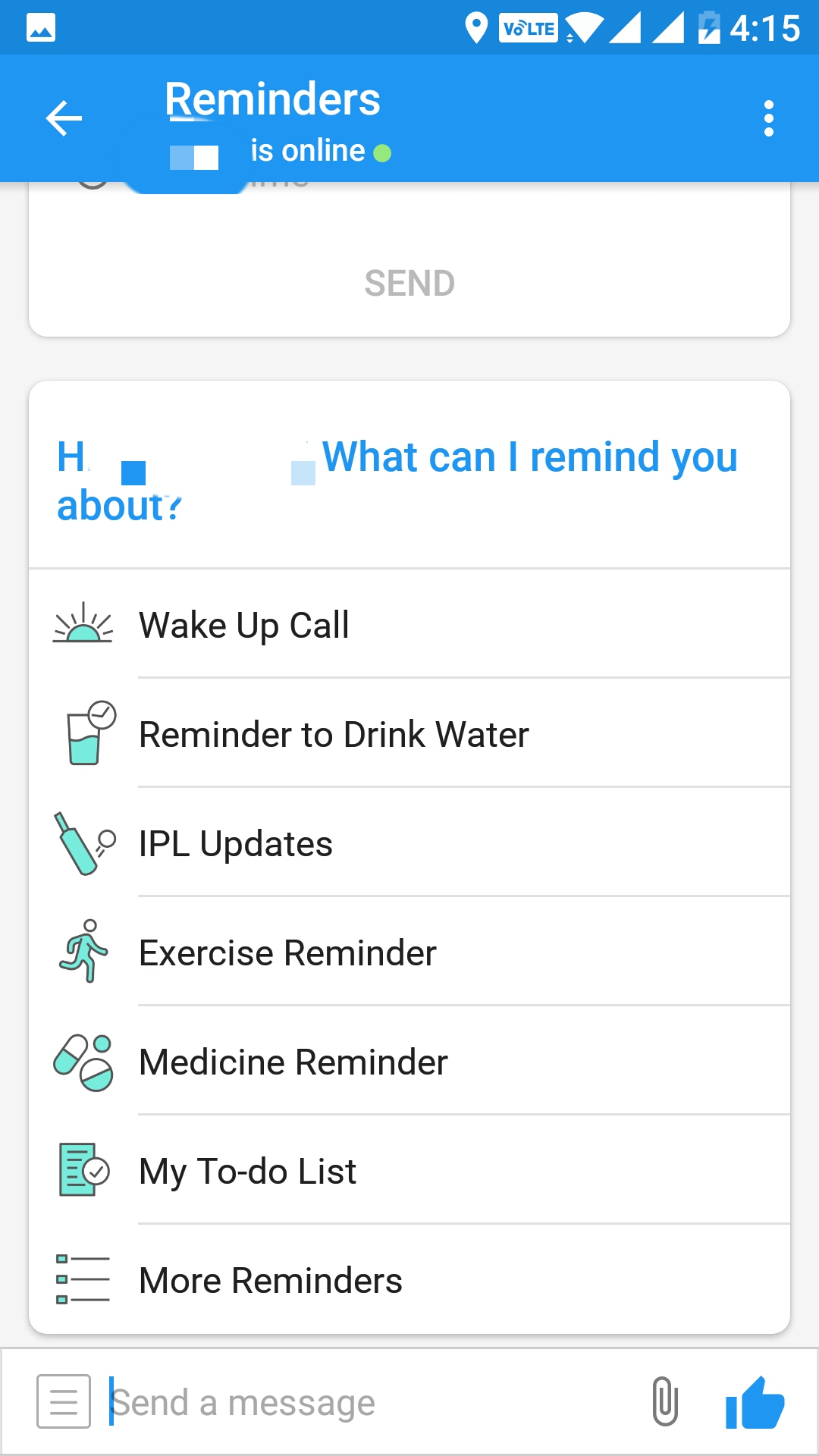}
  \caption{Reminders tasks}
  \label{fig:sfig2}
\end{subfigure}
\begin{subfigure}{.4\columnwidth}
  \centering
  \includegraphics[width=0.8\columnwidth]{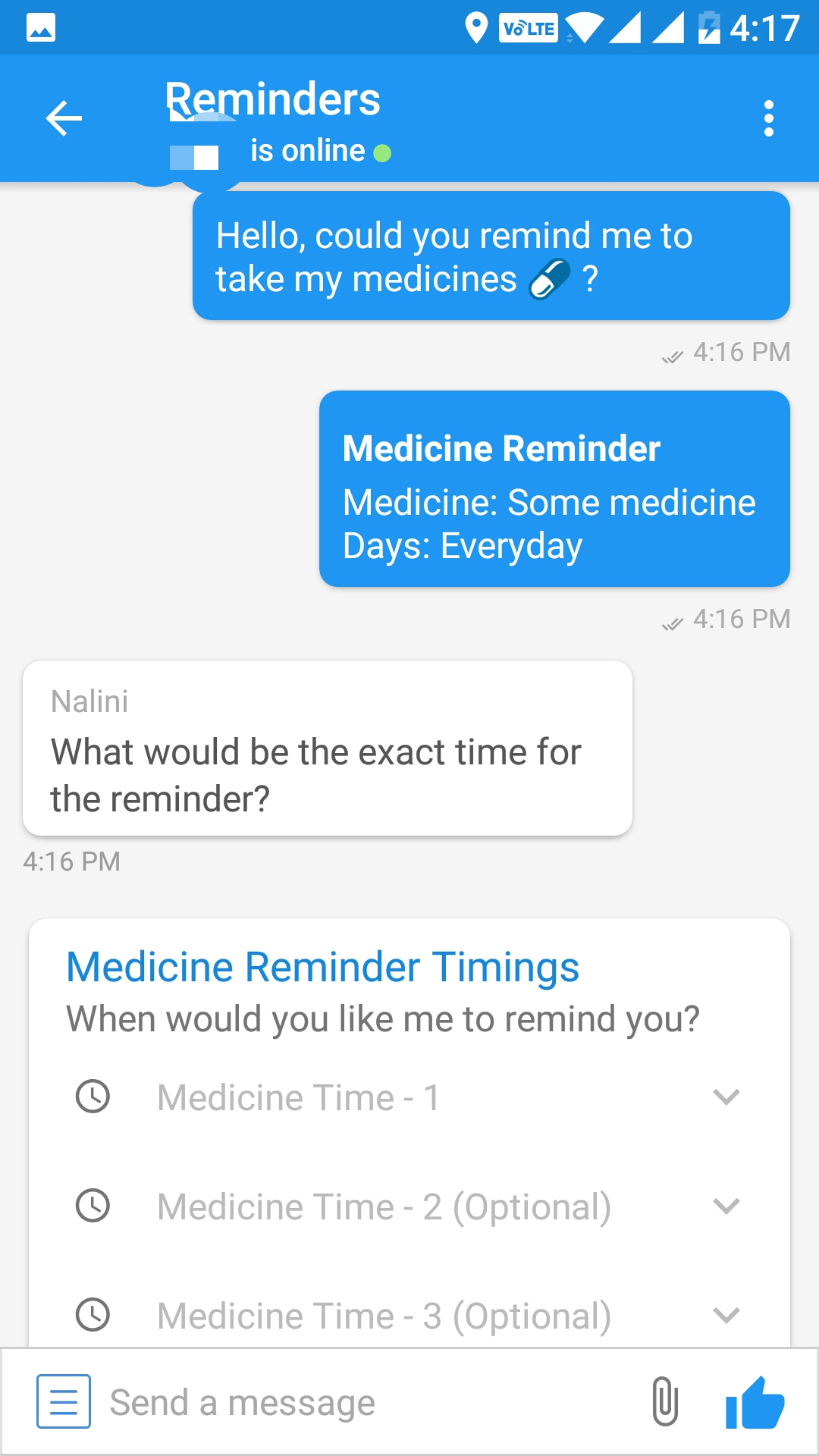}
  \caption{Reminder chat flow}
  \label{fig:sfig3}
\end{subfigure}
\caption{Overview of Haptik application on a smartphone. Screenshot \ref{fig:sfig2} lists down the tasks served by the Reminders service, \ref{fig:sfig3} shows a sample chat-flow for setting up a Medicine Reminder using UI elements such as a text-form}
\label{fig:haptik} 
\end{figure}

Our system is developed for Haptik\footnote{\url{https://haptik.ai}}, a personal assistant application that uses human agents to respond to user requests in the form of conversations. It caters to multiple services, ranging from scheduling reminders to planning travel itineraries. 
This paper focuses on automation of tasks involved in the Reminders Service. The tasks involved are scheduling reminders for specific tasks, notifying users with a message/call at a desired time and modifying the scheduled reminders. 
Figure \ref{fig:haptik} presents an overview of Reminders channel in Haptik application on a smartphone.
A typical conversation involves a user providing their intent and the chatbot trying to collect the required information like date, time, frequency and other contextual details like meeting venue. If the chatbot collects all the required information, it schedules a reminder. If at any point, the chatbot failed to address the query, the conversation would then be handed over to a human agent. Haptik provides UI elements aimed at assisting the user while conversing, like predictive response text giving the user an option to respond with either free-form natural language text or structured text.\\
With the hybrid system, we show that a retrieval model supported by a generative neural conversational model is much more robust than a simple rule assisted retrieval system. Our model is easier to scale compared to the existing models and is likely to offer improved results with more data thus making it a fit for production environments in the industry.
\section{Related Work}
\label{sect:relwork}
Two commonly used methods for responding to queries are Retrieval-based and Generative-based. Retrieval-based models use a repository of predefined responses and some heuristic to pick an appropriate response based on the input and context. The heuristic could be as simple as a rule-based expression match, or as complex as an ensemble of machine learning classifiers. These systems don’t generate new text, they only pick a response from a fixed set. \citeauthor{levin2000,young2010hidden,walker2003trainable,pieraccini2009we,wang2011improving} describe models that rely extensively on hand-coded rules, typically either building statistical models on top of heuristic rules or templates. Other models like those described by \citeauthor{oh2000stochastic,ratnaparkhi2002trainable,banchs2012iris,ameixa2014luke,nio2014developing,chen2013empirical} learn generation rules from a minimal set of authored rules.

Generative models build new responses from scratch. These models follow the line of investigation initiated by \citeauthor{ritter2011data} treating generation of conversational dialogue as a statistical machine translation problem. The SMT model proposed by Ritter et al., is end-to-end, purely data-driven, and contains no explicit model of dialog structure; the model learns to converse from human-to-human conversational corpora. \citeauthor{sordoni-EtAl:2015:NAACL-HLT} augments \citeauthor{ritter2011data} by re-scoring outputs using a seq2seq model \cite{sutskever2014sequence} conditioned on conversation history. Other researchers have recently used seq2seq to directly generate responses in an end-to-end fashion without relying on SMT phrase tables \cite{DBLP:conf/aaai/SerbanSBCP16,ShangLL15,DBLP:journals/corr/VinyalsL15}. \citeauthor{DBLP:conf/aaai/SerbanSBCP16} propose a hierarchical neural model aimed at capturing dependencies over an extended conversation history. Recent work by \citeauthor{li2016diversity} measures mutual information between message and response in order to reduce the proportion of generic responses typical of seq2seq systems. \citeauthor{DBLP:journals/corr/YaoZP15} employ an intention network to maintain the relevance of responses. Many systems learn to generate linguistically plausible responses, but they are not trained to generate semantically consistent ones because they are usually trained on a lot of data from multiple different users.

Most of the relevant works follow one of the above approaches. According to \citeauthor{wildml}, systems deployed at scale are more likely to be retrieval-based for now, because of low tolerance for errors as users expect humans to be responding. Both the approaches have some issues associated with them, we try a novel approach of combining the two models, to create a system which that addresses the individual issues of both the models.
\section{Graph Based Model}
\label{sect:graph}
We represent conversational flows as a graph that contains a set of nodes that represent a conversational state. The directional edges represent transition between the connected nodes. States represent certain steps or checkpoints in a task oriented dialog and actions are a set of defined ``actionables'' that the system can perform like calling third party APIs. The approach is similar to \cite{levin1997stochastic} that uses slot filling technique. The system uses heuristics to fill the slots and navigate from one state to another without any use of probability distribution. The reason to use such an approach is to build and modify the conversation flows when required without any use of prior conversational data. This approach doesn't make grammatical errors.

Each conversational state has unique intent linked with it, possible incoming user message templates and canned responses for the intent and set of actions to perform. For example, the objective of the state ``Wake up reminder'' is to set an alarm to wake up while the state ``Drink water reminder'' sets a reminder to drink water. States help keeping track of intent, updating entities, making API calls, etc. For e.g.  goal of the state ``set drink water reminder'' is to set a reminder to drink water by collecting/filling necessary entities/slots and calling the API to set the corresponding reminder. We introduce a connection between the two states with a directed edge, if one state is dependent on the other. This additional feature considers the previous state into account, while moving to the next state. The set of states and edges, forms a graph, defining some ideal chat-flows. The queries/intents and responses on a particular state are finite in number. Thus, when we traverse a graph, the traversal represents one of the ideal chat flows that we expect an user to follow. The states, actionables and the edges are handcrafted looking at the previous data and extracting the common chat-patterns followed by users.
\begin{figure}[ht]
  \includegraphics[width=0.47\textwidth]{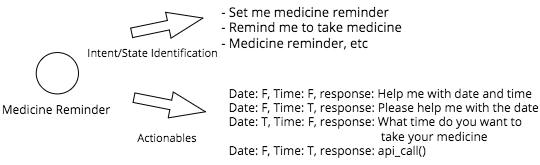}
  \caption{Medicine Reminder State with predefined set of templates}
\label{fig:graphsnippet1}
\end{figure}
\begin{figure*}[!htbp]
\begin{subfigure}{.25\textwidth}
  \centering
  \includegraphics[width=.7\linewidth]{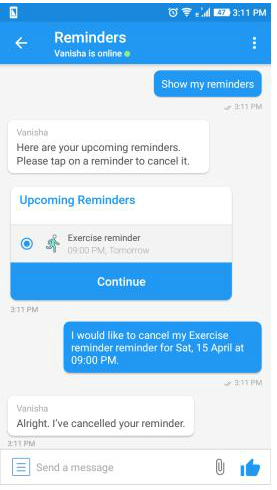}
  \caption{chat-flow}
  \label{fig:graphsfig1}
\end{subfigure}%
\begin{subfigure}{.7\textwidth}
  \centering
  \includegraphics[width=\linewidth]{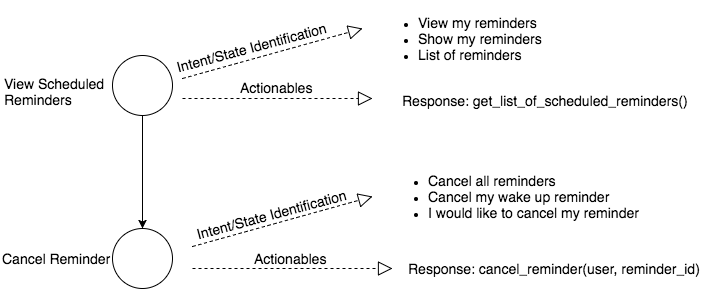}
  \caption{states, edges and actionables}
  \label{fig:graphsfig2}
\end{subfigure}
\caption{Reminder View and Reminder Cancel states connected with a directed edge} 
\label{fig:graph2}
\end{figure*}

For identifying a state we compute cosine similarity between the input message and each of the phrases associated with all possible next states after embedding them into the vector space using TF-IDF features and use the state that has the most similar phrase. Entity collection is performed using an in-house NER\footnote{\url{https://haptik.ai/chatbot-ner}}.
Consider, the snippet \ref{fig:graphsnippet1} of Medicine Reminder, where we have predefined set of templates that define the intent of the state and predefined set of slot filling table which map to responses to gather the necessary entity information. Chat-flow for the same can be seen in Figure 
\ref{fig:sfig3}. Consider another snippet 4 where an user is trying to view reminders and canceling it using UI elements. There are two states one to view reminders and other to cancel it.

While sophisticated retrieval based systems like a dual-encoder based models \cite{lowe2015ubuntu} have shown promising results on coverage of queries, we instead use the simple approach. Production systems undergo iterations to make changes. The reasons could be user/company demands, experimentation or to make user experience smoother. The changes could be addition or removal of features, addition of reminder types, short term campaigns or advertising campaigns.  Most of these changes require modifications to graph structure like addition or removal of states, edges, and actions. Our approach can adapt quickly to these changes with minimal amount of data in each state whereas a complex statistical approach might not adapt to the changes quickly with less data. While a classifier based approach is not a good fit for a production system like ours, we stick with the sentence similarity based approaches and keep on improving the technique for sentence matching.

\section{Neural Conversational Model}
\subsection{Model Description}
The neural conversational model described here is adapted from the seq2seq model described by \citeauthor{sutskever2014sequence} and modified for conversations by \citeauthor{DBLP:journals/corr/VinyalsL15}. A recurrent neural network (RNN) takes a sequence of words as input and predicts an output sequence. While training, conversations up to a point are fed into a RNN that encodes the information and produces a state or a thought vector. The thought vector is fed into the decoder which produces the output, token by token. We employ the use of the attention mechanism described in \cite{bahdanau2014neural} which lets the decoder selectively have a look at the input sequence, which helps overcome the limitation of a fixed context size regardless for long input sequences. In case of chat conversations, where there are multiple conversation turns, it is imperative to explore various semantic relations in input sequence while decoding the state.

Each input sequence is prepared after preprocessing each message and then concatenating multiple messages in a single conversation. Different messages from the same speaker are separated by a particular token while messages representing an end of turn i.e. switching of speaker between user and agent are represented by a separate token. To enforce a stricter limit on the length of the context, we restrict its input length to a certain number of words, prioritizing the most recent messages. The output messages are either text messages, keeping the graph state unchanged or suggest a transition to another node in the graph, for instance when the model suggests an API call to be made.

To tackle the problem of variable length sequences, we pad each input sequence and mark the end with a special token. The chat context can heavily differ in length and lead to the problem of sparse vectors, in case of very short input sequences. A bucketing approach is used where effectively multiple models are used which differ in the length of the input sequence and output sequence. Although these are separate models, they share the same parameters and hence are trained together. We use 3 layered bidirectional encoder with GRU \cite{cho2014properties} cells and 512 hidden units and unidirectional decoder with identical settings with attention over encoder states. We minimize the cross entropy loss calculated from the sampled softmax probability distribution over decoder outputs at each time step using SGD. We also employ recurrent dropout to avoid overfitting.
\subsection{Data}
We outline the data used for the Neural Conversation Model.
\subsubsection{Data Collection}
The corpus we use comprises of dyadic conversations from the Reminders domain. It is a collection of messages exchanged between Haptik users and chat assistants (responses from trained humans or the graph based model described in section \ref{sect:graph}). Participants of the dialog take alternate turns sending one or more messages per turn. Human assistants use a web interface to respond to user messages.
Due to limitations discussed later in this section, we use only most recently generated data (sample data\footnote{\url{https://haptik.ai/conversation-data}}) instead of the whole corpus. These include casual queries, out of domain requests, push notifications sent to the user.
But Reminders being a narrow domain, such a relatively smaller dataset still retains most requests we want to cater to.
The messages were observed to be mostly in English language but significant proportion of data also consists of code-mixed and code-switched messages - primarily English code-mixed with Hindi. Chat conversations offer acronyms and many contractions along with their short lengths. Each message has rich metadata accompanying it like user info (age, gender, location, device, etc.), timestamps and type of message (e.g. simple text, UI element, form, etc). The UI elements are represented in the Haptik specific format in the text corpus.
\subsubsection{Limitations}
As described earlier, production systems undergo iterations to add new functionalities as well as to update or stop supporting existing ones. These changes when made may alter the format for structured responses generated from non-textual UI elements or may completely change the way conversation flows. This introduces challenges in accumulating well varied data to train the model on, as data collected from recently introduced changes makes it available in significantly lesser proportions and variations relative to the whole dataset. 
\subsubsection{Preprocessing}
Before we train the seq2seq model, we perform several preprocessing steps, for efficient training with less data. With preprocessing we try to normalize the data, reduce vocabulary size, convert raw text into actionables, remove unnecessary data.
\begin{enumerate}
\item \textbf{Removal of out of domain conversations:}
Being a personal assistant app, users tend to ask queries from other domains in Reminders service. The first step involves removal of conversations from the data which do not belong to Reminders domain. 
\item \textbf{Merge Reminder Notifications:}
Users usually set multiple reminders on Haptik. For example, the drinking water reminder asks for how frequently should a user be reminded regarding the task. We often see a continuous series of reminder notifications in a conversation. We remove all such notifications except for the last one in the series. This reduces the number of outbound messages in data by a significant amount.
\item \textbf{Replace Entities:}
Using an in-house built NER, we replace the original text of the named entities such as date, time, phone number, user name and assistant name by placeholder tags like \textit{\_date\_}, \textit{\_time\_} and so on.

\item \textbf{Extracting actions from the message:}
This is a crucial part of the preprocessing step.The corpus consists of raw text messages exchanged between an user and a Haptik assistant. But being an utility product, just a text response is not enough to cater the user queries. We need to identify the action that needs to be performed. While scheduling or canceling a reminder, an acknowledgment message is sent to the user. In this step of preprocessing, we try to utilize these messages and tag them as an action. Whenever an action tag is predicted as a response by the neural model, we perform that action by fetching the corresponding state in the graph model, described in Section \ref{sect:graph}.\\
\textbf{E.g. Assistant Message:} Okay, done. We will remind you to take your medicine, via a call at 2:00 PM on Tue, 18 April. Take care :)\\
\textbf{Preprocessed message:} \textit{\_api\_call\_reminder\_medicine\_}\\
\textbf{State:} Medicine Reminder

\item\textbf{Orthography:}
We convert the string to lowercase and remove all the punctuation marks. 
At the end of this step, the data will contain characters only from the set \{a-z\} and a special character (`\_'). 
\end{enumerate}
\subsubsection{Data Statistics}
The dataset reduced from 10.8M to 3M messages after preprocessing. This is due to the removal of assistant messages consisting of push messages sent to begin a new conversation, merging of reminder notifications and removal of most out of domain conversations filtered out by the domain classifier, thus resulting in a decrease in the number of conversations as well. This is also supported by the balance introduced in the inbound and outbound message counts after preprocessing. It is observed that the mean number of messages exchanged dropped to half due to removal of lengthy casual conversations. Step 3 reduced the mean tokens per message to 6 owing to the replacement of multi word entities with a corresponding tag for the entity. Steps 3,5 resulted in the reduced vocabulary as expected. Finally, context-response pairs are generated from these 611K conversations with context being a concatenation of messages up to a point in the conversation(pruned to last 160 words) we want to encode and the immediate assistant message being the response we want the decoder to generate. We split these 659K pairs into training and validation sets in ratio 80:20.
\begin{table}[ht]
\begin{tabular}{|l|l|l|}
\hline
\textbf{Count of}                  & \textbf{Raw text}                & \textbf{\begin{tabular}[c]{@{}l@{}}After \\ pre-\\ processing\end{tabular}} \\ \hline
\textbf{Conversations} &1,093,443   &611,897\\ \hline
\textbf{Total messages}&10,887,922 &3,010,761\\ \hline
\textbf{Total Inbound messages} &1,675,428 &1,145,543\\ \hline
\textbf{Total Outbound messages} &9,212,492 & 1,865,218\\ \hline
\textbf{Messages per conversation*} &10 &5\\ \hline
\textbf{Turns per conversation*} &3 &3\\ \hline
\textbf{Tokens per message*} &14 &6\\ \hline
\textbf{Vocabulary} & $\approx$ 186,077   & $\approx$ 57,790\\ \hline
\textbf{Context-Target pairs} & NA & 659,542\\ \hline
\end{tabular}
\caption{Statistics about the data collected .Vocabulary consists of only alphabetic words converted to lowercase.* denotes mean over all counts.}
\label{tab:stats}
\end{table}
\section{Hybrid System}
\label{sect:hybrid}
The graph based system performs well for near-ideal scenarios. As per evaluation explained in section \ref{sect:resana}, we see that almost 70\% of the conversations are catered by graph based system. The rest 30\% of the chats fail due to either spelling mistakes, deviation from ideal chat-flows, code mixed queries or change of domain within the same conversation. To handle these issues, we introduce the Neural Conversation Model into the system. While training the Neural model we focus more on the queries from the rest 30\% conversations, as the graph model is not able to handle these queries.We train the model with 80\% human responses and 20\% graph responses. This trains the model to respond to queries where the graph model fails and also enables it to cover queries that belong to ideal flows which because of spell errors or localization are not served by the graph model. After building the neural conversation model we combine the two systems to run in parallel in real time. The flow diagram \ref{fig:hybrid} shows the working of the hybrid system. As expected, the hybrid system is able to handle the issues discussed above. Examples in tables \ref{tab:spell}, \ref{tab:daviation}, \ref{tab:domain} show how hybrid system help tackle the issues.
\subsection{Real time Working}
\begin{figure}
  \includegraphics[width=0.47\textwidth]{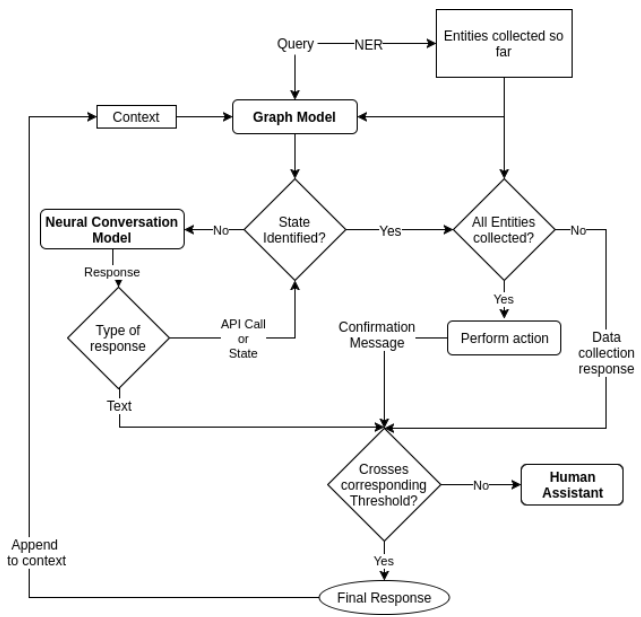}
  \caption{Working of the Hybrid System}
\label{fig:hybrid}
\end{figure}

Diagram \ref{fig:hybrid} provides an overview of the working of hybrid system in real time.
\textit{`Entities collected so far'}  and \textit{`Context'}, these two attributes represent all the events till that point in the conversation. For every query, we do some processing and update these two boxes with the new information. The \textit{`entities collected so far'} stores the domain specific (in this scenario: Reminders) entities identified in the queries asked so far (eg. date, time, name of medicine, etc.), whereas the \textit{`Context'} stores the messages exchanged so far. There is another object \textit{`state'} which stores the current graph based state of the chat. As explained earlier a state can have an associated action with it. Specifically in case of Reminders, every state has an associated action except for a special state, Generic State, which caters to most frequently asked common questions and also handles casual messages. When the graph based system fails to respond, neural system is requested to generate a response. If the neural engine emits an action, the action is performed and new state is updated to the state the action belongs to in the graph.

\begin{table}[b]
\centering
\begin{tabular}{@{}ll@{}}
\toprule
Sender & Message                                                                                                \\ \midrule
\textbf{User:}   & Hlo Ram                                                                                             \\
\textbf{Neural:} & Hi Shyam!                                                                                       \\
\textbf{User:}  & Can u plz wake me up                                                                                   \\
\textbf{Graph:}  & \begin{tabular}[t]{@{}l@{}}Sure Shyam, I will wake you up!\\  I just need a few details.\end{tabular} \\ \bottomrule
\end{tabular}
\caption{Spelling mistake handled}
\label{tab:spell}
\end{table}

\begin{table}[ht]
\centering
\begin{tabular}{ll}
\hline
\toprule
\textbf{User:}  & \begin{tabular}[t]{@{}l@{}}Can you call to another number\\  that I'll share now\end{tabular}                      \\ 
\textbf{Neural:} & \begin{tabular}[t]{@{}l@{}}I am sorry reminders can only be set\\  on the number registered with us.\end{tabular} \\ \hline
\end{tabular}
\caption{Deviation from ideal flow}
\label{tab:daviation}
\end{table}

\begin{table}[ht!]
\centering
\begin{tabular}{ll}
\hline
\textbf{User:}   & Hlooo                                                                                                                                                                                     \\ \cdashline{1-2}
\textbf{Neural:} & \begin{tabular}[t]{@{}l@{}}Hi John\\  how may I help you?\end{tabular}                                                                                                            \\ \cdashline{1-2}
\textbf{User:}  & \begin{tabular}[t]{@{}l@{}}Muje koi esi site ka\\  name btao jis se me\\  holywood movie downlod kr sku\\ (Tell me a website from where \\ I can download a Hollywood movie)\end{tabular} \\ \cdashline{1-2}
\textbf{Neural:} & \begin{tabular}[t]{@{}l@{}}Hum aapko link nahi send\\  kar sakta hu.\\ (I can not send a link to you)\end{tabular}                                                                        \\ \hline
\end{tabular}
\caption{Code mixing \& out of domain handled}
\label{tab:domain}
\end{table}
\subsection{Thresholding}
Incorrect responses in production systems are costly and may drive users away. We use two checks to stop the system from sending incorrect responses.
\begin{enumerate}
\item\textbf{Sentence Similarity Score:}
In Graph Model, states are identified based on phrases associated with it. The score is computed between the input query and the associated phrases. The state with the maximum score is considered if it exceeds a certain threshold.
\item\textbf{Heuristics on pipeline:}
The purpose of the Neural Model is to encourage user to follow one of the ideal chat-flows defined by the Graph Model, we shift the control whenever neural model fails to fulfill this purpose
\begin{itemize}
\item \textbf{Turns taken:} We shift the control of the chat if the number of continuous responses sent by the neural model exceeds a certain threshold. This case implies that the neural model failed to bring the user back to the chat-flow (as no response was sent by Graph Model).
\item \textbf{Responses not in favor of user:} We identify a generated response as \textit{`not favorable'} based on some heuristics. For example, if the response is ``Sorry I cannot help you with that''
. This type of response suggests that the response is not being able to cater the user efficiently. We restrict the neural model not to send such responses more than a threshold times in a conversation.
\end{itemize}
\end{enumerate}

\section{Results and Analysis }
\label{sect:resana}
\subsection{Evaluation Metric}
Recent works in response generation have adopted evaluation metrics from machine translation to compare a model’s generated response to a target response \citeauthor{liu2016not} show that these metrics correlate very weakly with human judgments. Production systems inherently provide supervised labels for evaluation, such as task completion. Our systems generate signals for task completion which can be used for evaluation. A conversation is marked as complete when a task is completed successfully ( the reminder is set, canceled, etc.) and is considered successfully handled if it was completed without any human intervention. We measure the performance of dialogue systems using percentage of chats that are completely handled by the dialogue system, without any kind of human intervention. We call this \textit{End to End Automation (E2E) Score}. There is a secondary evaluation metric, \textit{AOR Score}, which calculates percentage of chats in which at least one response is sent using automated dialogue system. For the purpose of analysis, we report the scores calculated over 5 days of real conversations. An end to end completed chat usually signifies that the user's query is successfully understood and the intended task is catered. The chats which are not end to end completed but at least one message is sent by dialogue system $(AOR-E2E)$ are usually the ones where initial messages from the user are catered while after some messages, either user's intent is not identified correctly or the task requested is out of scope of the system. 
\subsection{Results}
\begin{table}[ht]
\centering
\begin{tabular}{|l|l|l|l|}
\hline
\textbf{System} & \textbf{\begin{tabular}[c]{@{}l@{}}E2E\\ Score(\%)\end{tabular}} & \textbf{\begin{tabular}[c]{@{}l@{}}AOR\\ Score(\%)\end{tabular}} & \textbf{AOR-E2E} \\ \hline
\textbf{Graph} & 69.23 & 88.21 & 18.98 \\ \hline
\textbf{Hybrid} & 77.11 & 93.81 & 16.7 \\ \hline
\end{tabular}
\caption{Results (the scores are mean of 5 days of results)}
\label{tab:results}
\end{table}
The increment of 7.9\% in E2E score and 5.6\% in AOR score in the hybrid system clearly signifies that it outperforms graph model by handling its bottlenecks (i.e. spelling mistakes, deviation from ideal chat-flows, code mixed queries or out of domain queries) as mentioned in Section \ref{sect:hybrid}. The decrement in $AOR-E2E$ score signifies the conversion of at least 2\% of chats where the Graph System was earlier breaking after sending some messages, now being catered end to end by the the Hybrid system. Tables \ref{tab:spell} \ref{tab:daviation} \ref{tab:domain} are some of the examples where we find the Hybrid system improving the automation.
\subsection{Issues}
\label{ssect:issues}
Along with the improvements, new issues are introduced.
\begin{itemize}
\item\textbf{Domain Shift:} Users sometimes tend to start out of domain queries amidst of the conversation. Human agents are trained to cater to queries which belong to the domain of services offered by Haptik even when asked on Reminders channel. Such conversations are left unfiltered by the domain classifier as they started off with Reminders domain and thus make it into the training set. The model when tested generates a correct text response for such queries but fails to perform the correct actions because we did not tag out of domain actions during preprocessing.
\item\textbf{Gender Dependency:} Haptik assigns assistants randomly on the user facing application and thus our model was trained on responses by both male and female assistants. The replies generated by the model don't show consistency in using a particular gender. 
\item\textbf{Temporal Dependency:} The model has no sense of time while generating a response and often replies with the same ``Have a great day ahead'' or ``Good Morning'' greeting regardless the time of the day.
\item\textbf{Limited Vocabulary:} The model was trained on a relatively smaller vocabulary for a chatbot.
This leads to generation of related but incorrect responses. It might also happen that over time new words are introduced to the vocabulary. Training word vectors for a wider vocabulary can help addressing this issue. Recent works \cite{gu2016incorporating,gulcehre2016pointing,merity2016pointer} that employ copying mechanisms in seq2seq models can be used to mitigate this problem to some extent. For example:\\
\textit{User}: Happy Mahashivratri\\
\textit{Assistant (Model)}: Happy Diwali
\end{itemize}
\section{Conclusion and Future Work}
We have presented a hybrid approach to create robust production ready closed domain chatbots. We showed how a Generative Conversation Model can be used effectively to enhance the Retrieval Dialogue systems. Some of the issues like temporal and gender dependency could be solved using gender and time as features suring decoding. 
The graph based system easily adapts to the product changes. Next, we are focusing if the neural model can also be altered to learn the new changes with limited efforts. We performed some experiments, where we use the human data to continue training of the model. The major issue with this approach is to learn a new in-app feature. For e.g. when a new API call is introduced, we want to add a new tag (e.g. \textit{\_api\_modify\_reminder\_}) which does not exist already in the vocabulary. Instead of going through the entire training process again, during initial training we defined a vocabulary size which is more than the actual vocabulary size to act as buffer for new words in the future. Whenever new words are added, we assign them a location in the buffer space and with new training data, the embeddings for the tag are learned. With initial experiments, we found that the model is able to capture the new features with daily training.
\bibliography{aaai2018}

\begin{thebibliography}{}

\bibitem[\protect\citeauthoryear{Ameixa \bgroup et al\mbox.\egroup
  }{2014}]{ameixa2014luke}
Ameixa, D.; Coheur, L.; Fialho, P.; and Quaresma, P.
\newblock 2014.
\newblock Luke, i am your father: dealing with out-of-domain requests by using
  movies subtitles.
\newblock In {\em International Conference on Intelligent Virtual Agents},
  13--21.
\newblock Springer.

\bibitem[\protect\citeauthoryear{Bahdanau, Cho, and
  Bengio}{2014}]{bahdanau2014neural}
Bahdanau, D.; Cho, K.; and Bengio, Y.
\newblock 2014.
\newblock Neural machine translation by jointly learning to align and
  translate.
\newblock {\em arXiv preprint arXiv:1409.0473}.

\bibitem[\protect\citeauthoryear{Banchs and Li}{2012}]{banchs2012iris}
Banchs, R.~E., and Li, H.
\newblock 2012.
\newblock Iris: a chat-oriented dialogue system based on the vector space
  model.
\newblock In {\em Proceedings of the ACL 2012 System Demonstrations},  37--42.
\newblock Association for Computational Linguistics.

\bibitem[\protect\citeauthoryear{Britz}{2016}]{wildml}
Britz, D.
\newblock 2016.
\newblock Deep learning for chatbots, part 1 – introduction.

\bibitem[\protect\citeauthoryear{Chen, Wang, and
  Rudnicky}{2013}]{chen2013empirical}
Chen, Y.-N.; Wang, W.~Y.; and Rudnicky, A.~I.
\newblock 2013.
\newblock An empirical investigation of sparse log-linear models for improved
  dialogue act classification.
\newblock In {\em Acoustics, Speech and Signal Processing (ICASSP), 2013 IEEE
  International Conference on},  8317--8321.
\newblock IEEE.

\bibitem[\protect\citeauthoryear{Cho \bgroup et al\mbox.\egroup
  }{2014}]{cho2014properties}
Cho, K.; van Merri{\"e}nboer, B.; Bahdanau, D.; and Bengio, Y.
\newblock 2014.
\newblock On the properties of neural machine translation: Encoder--decoder
  approaches.
\newblock {\em Syntax, Semantics and Structure in Statistical Translation}
  103.

\bibitem[\protect\citeauthoryear{Gu \bgroup et al\mbox.\egroup
  }{2016}]{gu2016incorporating}
Gu, J.; Lu, Z.; Li, H.; and Li, V.~O.
\newblock 2016.
\newblock Incorporating copying mechanism in sequence-to-sequence learning.
\newblock {\em arXiv preprint arXiv:1603.06393}.

\bibitem[\protect\citeauthoryear{Gulcehre \bgroup et al\mbox.\egroup
  }{2016}]{gulcehre2016pointing}
Gulcehre, C.; Ahn, S.; Nallapati, R.; Zhou, B.; and Bengio, Y.
\newblock 2016.
\newblock Pointing the unknown words.
\newblock {\em arXiv preprint arXiv:1603.08148}.

\bibitem[\protect\citeauthoryear{Levin and
  Pieraccini}{1997}]{levin1997stochastic}
Levin, E., and Pieraccini, R.
\newblock 1997.
\newblock A stochastic model of computer-human interaction for learning
  dialogue strategies.
\newblock In {\em Eurospeech}, volume~97,  1883--1886.

\bibitem[\protect\citeauthoryear{Levin, Pieraccini, and
  Eckert}{2000}]{levin2000}
Levin, E.; Pieraccini, R.; and Eckert, W.
\newblock 2000.
\newblock A stochastic model of human-machine interaction for learning dialog
  strategies.
\newblock {\em {IEEE} Trans. Speech and Audio Processing} 8(1):11--23.

\bibitem[\protect\citeauthoryear{Li \bgroup et al\mbox.\egroup
  }{2016}]{li2016diversity}
Li, J.; Galley, M.; Brockett, C.; Gao, J.; and Dolan, B.
\newblock 2016.
\newblock A diversity-promoting objective function for neural conversation
  models.
\newblock In {\em Proceedings of NAACL-HLT},  110--119.

\bibitem[\protect\citeauthoryear{Liu \bgroup et al\mbox.\egroup
  }{2016}]{liu2016not}
Liu, C.-W.; Lowe, R.; Serban, I.~V.; Noseworthy, M.; Charlin, L.; and Pineau,
  J.
\newblock 2016.
\newblock How not to evaluate your dialogue system: An empirical study of
  unsupervised evaluation metrics for dialogue response generation.
\newblock {\em arXiv preprint arXiv:1603.08023}.

\bibitem[\protect\citeauthoryear{Lowe \bgroup et al\mbox.\egroup
  }{2015}]{lowe2015ubuntu}
Lowe, R.; Pow, N.; Serban, I.; and Pineau, J.
\newblock 2015.
\newblock The ubuntu dialogue corpus: A large dataset for research in
  unstructured multi-turn dialogue systems.
\newblock {\em arXiv preprint arXiv:1506.08909}.

\bibitem[\protect\citeauthoryear{Merity \bgroup et al\mbox.\egroup
  }{2016}]{merity2016pointer}
Merity, S.; Xiong, C.; Bradbury, J.; and Socher, R.
\newblock 2016.
\newblock Pointer sentinel mixture models.
\newblock {\em arXiv preprint arXiv:1609.07843}.

\bibitem[\protect\citeauthoryear{Nio \bgroup et al\mbox.\egroup
  }{2014}]{nio2014developing}
Nio, L.; Sakti, S.; Neubig, G.; Toda, T.; Adriani, M.; and Nakamura, S.
\newblock 2014.
\newblock Developing non-goal dialog system based on examples of drama
  television.
\newblock In {\em Natural Interaction with Robots, Knowbots and Smartphones}.
  Springer.
\newblock  355--361.

\bibitem[\protect\citeauthoryear{Oh and Rudnicky}{2000}]{oh2000stochastic}
Oh, A.~H., and Rudnicky, A.~I.
\newblock 2000.
\newblock Stochastic language generation for spoken dialogue systems.
\newblock In {\em Proceedings of the 2000 ANLP/NAACL Workshop on Conversational
  systems-Volume 3},  27--32.
\newblock Association for Computational Linguistics.

\bibitem[\protect\citeauthoryear{Pieraccini \bgroup et al\mbox.\egroup
  }{2009}]{pieraccini2009we}
Pieraccini, R.; Suendermann, D.; Dayanidhi, K.; and Liscombe, J.
\newblock 2009.
\newblock Are we there yet? research in commercial spoken dialog systems.
\newblock In {\em International Conference on Text, Speech and Dialogue},
  3--13.
\newblock Springer.

\bibitem[\protect\citeauthoryear{RATNAPARKHI}{2002}]{ratnaparkhi2002trainable}
RATNAPARKHI, A.
\newblock 2002.
\newblock Trainable approaches to surface natural language generation and their
  application to conversational dialog systems.
\newblock {\em Computer speech \& language} 16(3-4):435--455.

\bibitem[\protect\citeauthoryear{Ritter, Cherry, and
  Dolan}{2011}]{ritter2011data}
Ritter, A.; Cherry, C.; and Dolan, W.~B.
\newblock 2011.
\newblock Data-driven response generation in social media.
\newblock In {\em Proceedings of the conference on empirical methods in natural
  language processing},  583--593.
\newblock Association for Computational Linguistics.

\bibitem[\protect\citeauthoryear{Serban \bgroup et al\mbox.\egroup
  }{2016}]{DBLP:conf/aaai/SerbanSBCP16}
Serban, I.~V.; Sordoni, A.; Bengio, Y.; Courville, A.~C.; and Pineau, J.
\newblock 2016.
\newblock Building end-to-end dialogue systems using generative hierarchical
  neural network models.
\newblock In {\em Proceedings of the Thirtieth {AAAI} Conference on Artificial
  Intelligence, February 12-17, 2016, Phoenix, Arizona, {USA.}},  3776--3784.

\bibitem[\protect\citeauthoryear{Shang, Lu, and Li}{2015}]{ShangLL15}
Shang, L.; Lu, Z.; and Li, H.
\newblock 2015.
\newblock Neural responding machine for short-text conversation.
\newblock  1577--1586.
\newblock The Association for Computer Linguistics.

\bibitem[\protect\citeauthoryear{Sordoni \bgroup et al\mbox.\egroup
  }{2015}]{sordoni-EtAl:2015:NAACL-HLT}
Sordoni, A.; Galley, M.; Auli, M.; Brockett, C.; Ji, Y.; Mitchell, M.; Nie,
  J.-Y.; Gao, J.; and Dolan, B.
\newblock 2015.
\newblock A neural network approach to context-sensitive generation of
  conversational responses.
\newblock In {\em Proceedings of the 2015 Conference of the North American
  Chapter of the Association for Computational Linguistics: Human Language
  Technologies},  196--205.
\newblock Denver, Colorado: Association for Computational Linguistics.

\bibitem[\protect\citeauthoryear{Sutskever, Vinyals, and
  Le}{2014}]{sutskever2014sequence}
Sutskever, I.; Vinyals, O.; and Le, Q.~V.
\newblock 2014.
\newblock Sequence to sequence learning with neural networks.
\newblock In {\em Advances in neural information processing systems},
  3104--3112.

\bibitem[\protect\citeauthoryear{Vinyals and
  Le}{2015}]{DBLP:journals/corr/VinyalsL15}
Vinyals, O., and Le, Q.~V.
\newblock 2015.
\newblock A neural conversational model.
\newblock {\em CoRR} abs/1506.05869.

\bibitem[\protect\citeauthoryear{Walker, Prasad, and
  Stent}{2003}]{walker2003trainable}
Walker, M.; Prasad, R.; and Stent, A.
\newblock 2003.
\newblock A trainable generator for recommendations in multimodal dialog.
\newblock In {\em In Proc. EUROSPEECH}.

\bibitem[\protect\citeauthoryear{Wallace}{2003}]{wallace2003elements}
Wallace, R.
\newblock 2003.
\newblock The elements of aiml style.
\newblock {\em Alice AI Foundation}.

\bibitem[\protect\citeauthoryear{Wang \bgroup et al\mbox.\egroup
  }{2011}]{wang2011improving}
Wang, W.~Y.; Artstein, R.; Leuski, A.; and Traum, D.~R.
\newblock 2011.
\newblock Improving spoken dialogue understanding using phonetic mixture
  models.
\newblock In {\em FLAIRS Conference}.

\bibitem[\protect\citeauthoryear{Yao, Zweig, and
  Peng}{2015}]{DBLP:journals/corr/YaoZP15}
Yao, K.; Zweig, G.; and Peng, B.
\newblock 2015.
\newblock Attention with intention for a neural network conversation model.
\newblock {\em CoRR} abs/1510.08565.

\bibitem[\protect\citeauthoryear{YOUNG}{2010}]{young2010hidden}
YOUNG, S.
\newblock 2010.
\newblock The hidden information state model: A practical framework for
  pomdp-based spoken dialogue management.
\newblock {\em Comput. Speech Lang.} 24:150--174.

\end{thebibliography}
\bibliographystyle{aaai}
\end{document}